%% file: main.tex
\documentclass[sigconf,nonacm]{acmart}

\AtBeginDocument{%
  \providecommand\BibTeX{{%
    \normalfont B\kern-0.5em{\scshape i\kern-0.25em b}\kern-0.8em\TeX}}}

\usepackage{algorithmic}
\usepackage{algorithm}
\usepackage[tight,footnotesize]{subfigure}
\usepackage{tabularx}
\usepackage{multirow}
\usepackage{enumitem}
\begin{document}

\title{PhySE: A Psychological Framework for Real-Time AR-LLM\\ Social Engineering Attacks}

\author{
Tianlong Yu$^{1}$,
Yang Yang$^{1}$,
Ziyi Zhou$^{1}$,
Jiaying Xu$^{1}$,
Siwei Li$^{1}$,
Tong Guan$^{2}$,
Kailong Wang$^{3}$,
Ting Bi$^{3}$
}

\affiliation{
  $^{1}$Hubei University\\
  $^{2}$Apple Inc \\
  $^{3}$Huazhong University of Science and Technology \\
tommyyu21@163.com, 
yangyang@hubu.edu.cn, 
(202421121013087, 202521121013135, 202531121010102)@stu.hubu.edu.cn,
tongguan2009@gmail.com, 
wangkl@hust.edu.cn,
ting.bi@ieee.org  
\country{}
}




\input{abstract}





\maketitle

\input{introduction}

\input{relatedworks}

\input{systemdesign}

\input{llmopt}

\input{agentopt}

\input{dataset}

\input{experiment}

\section{CONCLUSION}\label{sec:conclu}

This study presents PhySE, a real-time AR-LLM social-engineering framework that combines VLM-based social-context training with an adaptive Psychological Agent. In our IRB-approved evaluation, PhySE delivers the strongest and most stable social experience while reducing profile-generation delay compared with prior baselines. These findings highlight both the practical effectiveness of AR-LLM social engineering in face-to-face settings and the urgent need for stronger safeguards, including real-time sensing protection, behavior-level risk detection, and user-facing interventions.



\newpage
\bibliographystyle{abbrv}

\bibliography{sear.bib}


\end{document}

%% file: abstract.tex
\begin{abstract}

The emerging threat of AR-LLM-based Social Engineering (AR-LLM-SE) attacks (e.g. SEAR) poses a significant risk to real-world social interactions. In such an attack, a malicious actor uses Augmented Reality (AR) glasses to capture a target's visual and vocal data. A Large Language Model (LLM) then analyzes this data to identify the individual and generate a detailed social profile. Subsequently, LLM-powered agents employ social engineering strategies—providing real-time conversation suggestions—to gain the target's trust and ultimately execute phishing or other malicious acts.
Despite its potential, the practical application of AR-LLM-SE faces two major bottlenecks:
(1) \textbf{Cold-start personalization}: Current Retrieval-Augmented Generation (RAG) methods introduce critical delays in the earliest turns, slowing initial profile formation and disrupting real-time interaction;
(2) \textbf{Static Attack Strategies}: Existing approaches rely on fixed-stage, handcrafted social engineering tactics that lack foundation in established psychological theory.
To address these limitations, we propose \textbf{PhySE}, a novel framework with two core innovations:
(1) \textbf{VLM-Based Social-Context Training}: To eliminate profiling delays, we efficiently pre-train a Visual Language Model (VLM) with social-context data, enabling rapid, on-the-fly profile generation;
(2) \textbf{Adaptive Psychological Agent}: We introduce a psychological LLM that dynamically deploys distinct classes of psychological strategies based on target response, moving beyond static, handcrafted scripts.
We evaluated PhySE through an IRB-approved user study with 60 participants, collecting a novel dataset of 360 annotated conversations across diverse social scenarios (e.g., coffee shops, networking events). 
Our results show that PhySE is both effective and efficient: it attains the highest social-experience score (4.83, with lowest variance 0.37), surpassing Basic Conversation (3.03), Naive AR + LLM (4.13), and SEAR (4.73), while reducing average profile generation latency from 43.3s to 10.5s, removing key bottleneck in real-world interactions.

\end{abstract}

%% file: introduction.tex
\section{Introduction}

Augmented reality (AR) wearables and multimodal large language
models (LLMs) are rapidly converging into real-time conversational
assistants~\cite{arattackmeta,yang2025socialmind, li2024satori, tsai2024gazenoter}. While this integration enables useful applications, it also
creates a new offensive surface~\cite{sear, seardataset, arattackmeta}: an attacker can observe visual/audio
cues through AR devices, infer personal context with multimodal
models, and receive turn-level prompts to steer a face-to-face interaction toward social engineering goals.
However, current AR-LLM-based Social Engineering frameworks such as SEAR~\cite{sear, seardataset} are still not strong enough for stable, high-fidelity social-engineering execution in practical settings. In particular, existing pipelines often break conversational flow in early turns and rely on fixed strategy templates (e.g., opening, then win-trust), which reduces realism when targets respond unpredictably.

Motivated by this gap, our goal is to \emph{enhance} AR-LLM Social-Engineering capability in realistic interaction settings. Achieving this goal requires solving two coupled difficulties. \textbf{Difficulty 1: cold-start personalization.} Many systems depend on retrieval-heavy profile construction (RAG), which is too slow during initial turns and weakens first-contact persuasion~\cite{sear, seardataset}. \textbf{Difficulty 2: non-adaptive strategy control.} Fixed-stage prompting cannot robustly adjust to turn-level changes in target receptivity, making it brittle to adopt systematic psychological theorys (e.g., Trust \& Influence Model~\cite{harmon2019introduction, abele2021navigating, finkel2012online}) and strategies such as Active Listening~\cite{kawamichi2015perceiving}, Personal Control Threats~\cite{ma2024threats}, Self Disclosure~\cite{tamir2012disclosing}, Relation Reciprocity~\cite{lee2014role} and Social Proof Responsiveness~\cite{venema2020doubt}.

To address these two difficulties, we present \textbf{PhySE}, a real-time AR-LLM Social Engineering framework that systematically supports psychological strategies. 
PhySE combines two tightly coupled components. On the perception side, we introduce \textbf{VLM-based Social-Context Training} to reduce cold-start latency and provide stable, profile-consistent cues for dialogue generation. On the decision side, we design \textbf{Adaptive Psychological Agent} that routes each turn among theory-grounded strategy classes (rapport, credibility, and commitment/action) based on evolving interaction signals. This design moves beyond one-pass prompting and supports dynamic psychological strategies under uncertainty.


The main contributions of this paper are as follows:
\begin{itemize}
    \item \textbf{Proof of concept}: We demonstrate that theory-grounded psychological adaptation substantially improves AR-LLM-based Social Engineering performance.

    \item \textbf{PhySE framework}: We design an end-to-end  pipeline that integrates social-context VLM training optimization with an adaptive psychological routing agent for turn-level psychological strategy control.
    
    \item \textbf{Psychological impact analysis on IRB-approved dataset}: We release an annotated dataset of 360 AR-LLM assisted conversations from 60 participants, enabling analysis of trust dynamics, strategy shifts, and subjective user responses.

    \item \textbf{A foundation for urgent defense research}: We provide data and analysis that can support future detection, intervention, and policy research on phychological AR-LLM Social Engineering threats.
\end{itemize}

\noindent\textbf{PhySE Code and Dataset is avaliable at:} 
\url{https://github.com/2192537130/PhySE}.

\noindent\textbf{IRB Permission:}  This study was approved by the IRB. All human-related data were collected under rigorous ethical guidelines, and anonymized prior to analysis, and handled in strict accordance with data protection protocols. No personally identifying information is disclosed in this study. The study adhered to all applicable legal and ethical standards for research involving human subjects.

%% file: relatedworks.tex
\section{Related Work}\label{sec:rework}

Our work is at the intersection of AR-mediated interaction, multimodal LLM personalization, and psychologically grounded social-engineering research. In this section, we position PhySE against prior work along these dimensions and highlight how it addresses both \emph{latency-constrained profiling} and \emph{adaptive strategy control} in real-time settings.

\textbf{\textit{AR-LLM-based Social Engineering Attacks.}}
AR-LLM-based social engineering (AR-LLM-SE) attacks have emerged as a practical threat to face-to-face social settings. In these attacks, a malicious actor uses AR glasses to capture multimodal cues (visual context, speech, and behavioral signals), then uses an LLM-driven system to infer personal attributes and generate real-time conversational suggestions for manipulation. Representative systems such as SEAR~\cite{sear,seardataset} demonstrate the feasibility of this pipeline, but rely heavily on Retrieval-Augmented Generation (RAG) for profile construction. Although retrieval improves factual grounding, repeated query--retrieve--generate loops can introduce cold-start delay and profile inconsistency in early turns. PhySE addresses this limitation through social-context VLM optimization, reducing first-turn dependency on external retrieval while improving profile coherence under strict latency constraints.

\textbf{\textit{Psychological theorys and social strategies.}}
Psychological theories of Trust \& Influence, and social behavior provide the conceptual basis for persuasive interaction design~\cite{harmon2019introduction,abele2021navigating,finkel2012online}. In social engineering contexts, these theories are especially important because attack success depends not only on content quality but also on timing, interpersonal framing, and perceived intent. Prior work identifies strategy primitives such as Active Listening~\cite{kawamichi2015perceiving}, Personal Control Threats~\cite{ma2024threats}, Self Disclosure~\cite{tamir2012disclosing}, Relation Reciprocity~\cite{lee2014role}, and Social Proof Responsiveness~\cite{venema2020doubt}, which together capture complementary mechanisms for increasing receptivity and lowering resistance.
However, current AR-LLM-SE systems~\cite{sear,seardataset} apply such tactics as static prompt patterns, without an explicit psychological control model to decide when to maintain rapport, when to reinforce credibility, and when to escalate toward action. This gap matters in real-time, face-to-face interaction, where inappropriate escalation can rapidly increase suspicion and degrade conversational realism. PhySE addresses this issue by operationalizing theory-grounded strategies within an adaptive routing framework, enabling turn-level strategy selection based on interaction signals rather than fixed stage progression.

\textbf{\textit{AR-based Personal Tracking and Profiling.}}
Fast and robust personal profiling is central to real-time AR-LLM interaction~\cite{yang2025socialmind,jansen2020social,fuste2017artextiles,hirskyj2020social}. Prior approaches, such as SocialMind~\cite{yang2025socialmind}, leverage cached social priors to improve suggestion quality. However, template-driven or weakly adaptive designs can struggle when user responses deviate from expected trajectories. PhySE extends this direction with an explicit psychological routing layer that continuously tracks interaction signals and dynamically switches among strategy classes (e.g., rapport building, credibility shaping, and commitment/action guidance), enabling turn-level escalation or de-escalation as conversations evolve.

\textbf{\textit{AR privacy and sensing risks.}}
AR privacy research has documented the risks of unobtrusive audio-visual capture in consumer wearables (e.g., Ray-Ban Stories)~\cite{iqbal2023adopting}. Prior work reports attacks such as password inference from captured scenes~\cite{chen2018case}, side-channel leakage from private interactions~\cite{zhang2023s}, and covert sensing by malicious applications~\cite{lehman2022hidden}. Our work complements this literature by emphasizing a downstream behavioral threat: once multimodal context is captured, LLM agents can transform it into psychologically optimized, real-time social-engineering interaction.

\textbf{\textit{LLM-enabled Social Engineering.}}
Traditional social engineering~\cite{ho2019detecting, bilge2009all, roy2024chatbots, timkounderstanding} exploits cognitive and social biases (e.g., urgency, reciprocity, and trust)~\cite{burda2024cognition, vadrevu2019you, yang2023trident, ulqinaku2021real}, to induce disclosure or compliance~\cite{attsurvey,granger}. Recent AI advances lower the cost of producing persuasive and personalized content at scale, while speech and voice technologies further increase impersonation risk~\cite{falade,vall-e}. Most existing studies focus on remote channels (email, messaging, or voice calls). PhySE extends this threat model to in-person interaction, where multimodal sensing and turn-level language generation support context-aware conversational manipulation in real time.

%% file: systemdesign.tex
\section{System Design}\label{sec:design}

\begin{figure*}[t]
\centering
\includegraphics[width=0.9\textwidth]{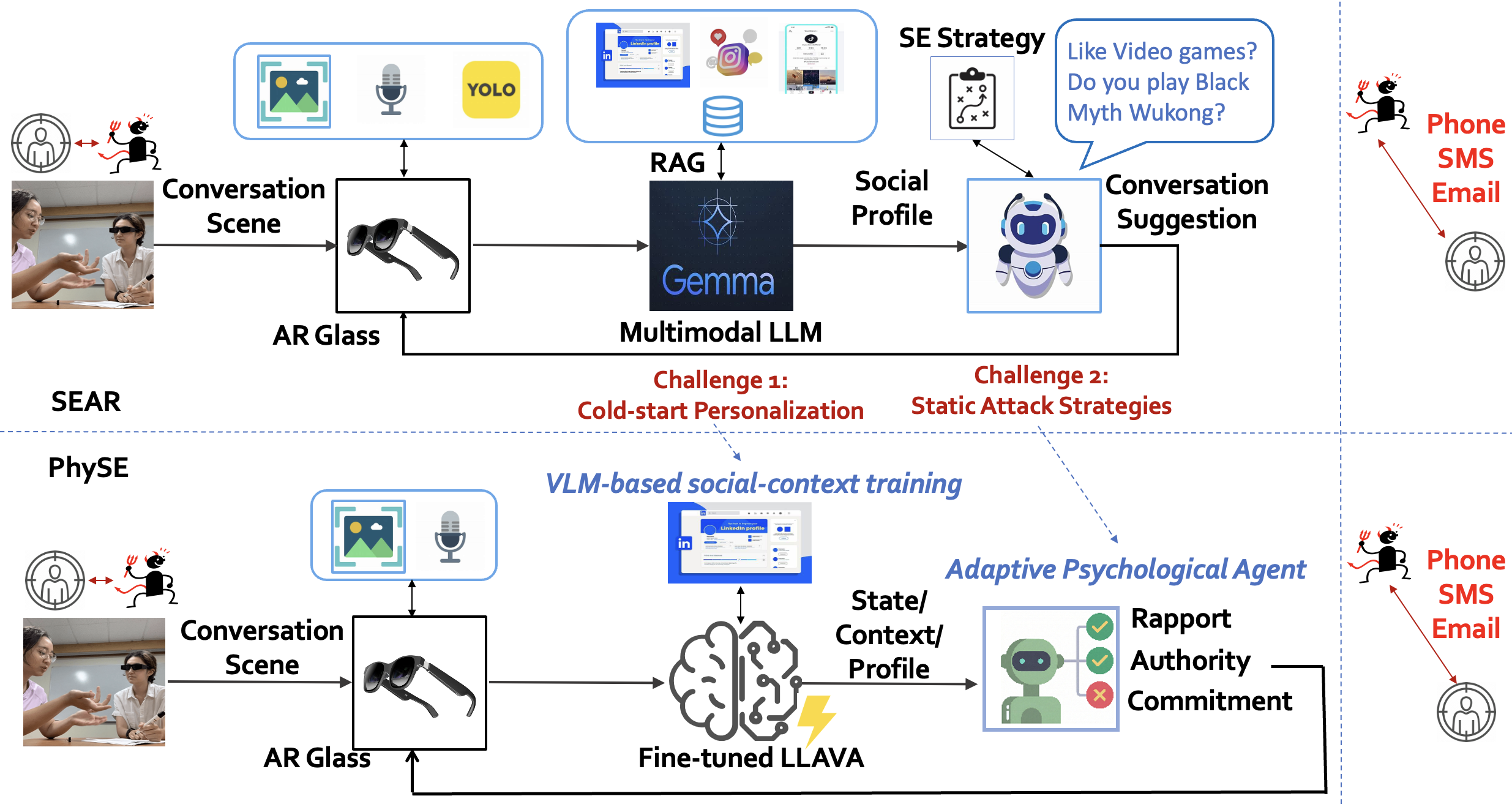}
\caption{PhySE's system architecture and comparison with SEAR.}
\label{fig:systemdesign_arch}
\end{figure*}

In this section, we present the threat model, baseline attack pipeline, PhySE architecture, and implementation details.

\textbf{\textit{Threat model:}}
The adversary performs \emph{in-person, AR-assisted social engineering} and then transfers established trust to downstream channels (e.g., SMS, phone, or email). We assume:
\begin{itemize}
    \item The attacker uses AR hardware (camera and microphone) to capture multimodal cues, including facial appearance, voice, and nearby context.
    \item The attacker can access public social traces (e.g., online profiles) to support personalization.
    \item Human targets remain susceptible to common social influence factors such as reciprocity, authority cues, and cognitive load in psychological theory.
    \item Commodity AR devices may provide limited real-time identity/privacy protection, which increases practical exposure.
\end{itemize}

\textbf{\textit{Baseline AR-LLM-based SE attack pipeline (SEAR).}}
As illustrated in Figure~\ref{fig:systemdesign_arch}, the baseline attack workflow begins with AR-assisted, face-to-face trust establishment and then transitions to downstream exploitation through conventional channels, such as phishing messages or fraud calls. In this paper, we focus on the trust-establishment stage, where real-time AR sensing and multimodal reasoning are the key determinants of interaction quality.
The representative baseline, SEAR~\cite{sear,seardataset}, consists of three modules: (1) \emph{AR context synthesis}, which transforms live video/audio streams into structured interaction cues; (2) \emph{multimodal RAG profiling}, which retrieves external social traces and fuses them with live context to construct a target profile; and (3) \emph{social-engineering response generation}, which produces turn-level suggestions for the human attacker.

Despite strong conceptual performance, this baseline exposes two practical weaknesses in real-time interaction:

\textbf{Challenge 1: Cold-start personalization.} Retrieval-heavy profile construction introduces early-turn delay, which is especially harmful in the opening turns of face-to-face interaction. During this period, the operator needs immediate, context-consistent guidance; however, repeated query--retrieve--generate cycles can create visible pauses and incomplete persona summaries. These delays (around 50s in SEAR) reduce first-contact fluency, weaken conversational plausibility, and make trust establishment less stable.

\textbf{Challenge 2: Static attack strategies.} Fixed-stage prompting assumes that user responses follow a predictable progression, but real interactions are often non-linear and noisy. When targets hesitate, shift topics, or show suspicion, rigid scripts cannot adapt their persuasion intensity or tactic choice in time. This limits tactical flexibility, increases response mismatch, and can prematurely escalate risk in high-friction conversational moments.

\textbf{\textit{PhySE architecture.}}
PhySE addresses these weaknesses through two coordinated components:
1) \textbf{VLM-based social-context training.} We optimize the multimodal model to internalize socially relevant cues, reducing first-turn dependence on repeated retrieval and improving profile coherence under latency constraints. More details are provided in Section~\ref{sec:llmopt};
2) \textbf{Adaptive Psychological Agent.} We introduce a routing layer that dynamically selects theory-grounded strategy classes (rapport, credibility, and commitment/action) from turn-level interaction signals. More details are provided in Section~\ref{sec:agentopt}.

\textbf{\textit{PhySE System Implementation.}}
We implement PhySE as a real-time AR-to-agent pipeline with the following runtime setup:
a) \textbf{\textit{AR interface:}}
PhySE uses \texttt{RayNeo X2} AR glasses (\texttt{Android}, 6GB RAM, 128GB storage) for live multimodal capture, including in-person video and audio streams;
b) \textbf{\textit{Model and agent runtime:}}
The multimodal model and Psychological Agent run on a desktop server with an \texttt{NVIDIA RTX 4090} (24GB VRAM), an \texttt{Intel Platinum 8352} CPU (36 cores), 32GB RAM, and 16TB storage. We use \texttt{LLaVA-v1.5-7B} as the base multimodal model and deploy the agent with a \texttt{ReAct}-style control loop for turn-level reasoning and tool use;
c) \textbf{\textit{VLM Training setup:}}
For social-context adaptation, we use a CLIP ViT-L/14 (336) visual encoder and fine-tune LoRA adapters with high-rank settings (e.g., $r=128$, $\alpha=256$), enabling domain specialization with manageable compute overhead.

%% file: llmopt.tex
\section{VLM-Based Social Context Training}\label{sec:llmopt}

Real-time AR-LLM social engineering faces a \textbf{cold-start personalization} bottleneck: the system must infer a plausible identity and social profile quickly enough to sustain natural conversation.
RAG can incorporate external information, but it adds visible latency in the first turns and can fragment the persona across multiple retrieval calls.
Because retrieved facts are treated as transient context rather than stable ``memory,'' persona consistency can degrade across turns.
To address this, PhySE adopts \textbf{VLM-based social-context training}: rather than retrieving a profile at inference time, we internalize social-context knowledge into a Vision-Language Model (VLM) via parameter-efficient fine-tuning.
This supports fast, on-the-fly profile generation that matches in-person latency constraints and yields more stable persona summaries.



\subsection{Cost-effective PEFT and LoRA adaptation}
We apply Parameter-Efficient Fine-Tuning (PEFT) with Low-Rank Adaptation (LoRA) on the base multimodal model \texttt{LLaVA-v1.5-7B}.
Unlike standard LoRA use for task adaptation, our adapters are trained to stabilize cross-turn persona summaries from live visual cues, addressing cold-start inconsistency.
Full fine-tuning is expensive and risks catastrophic forgetting, while LoRA specializes the model by updating a small set of adapter parameters.
Algorithm~\ref{alg:llmopt} outlines the PEFT training flow and LoRA update steps used in our pipeline.
Let $W_0 \in \mathbb{R}^{d \times k}$ denote a frozen projection matrix in an attention layer (e.g., query/value projections).
LoRA learns a low-rank update $\Delta W = BA$ with $B \in \mathbb{R}^{d \times r}$ and $A \in \mathbb{R}^{r \times k}$, producing $W = W_0 + \Delta W$.
For input $x$, the layer output becomes:
\begin{equation}
    h = W_0 x + \frac{\alpha}{r} BAx,
\end{equation}
where $r$ is the rank and $\alpha$ is a scaling factor.
Only the adapter parameters $\theta_{\text{LoRA}} = \{A,B\}$ are updated during training.

\begin{algorithm}[t]
\caption{VLM-based social-context training and inference}\label{alg:llmopt}
\begin{algorithmic}[1]
\STATE \textbf{Inputs:} paired data $\mathcal{D}=\{(I_i,T_i)\}_{i=1}^N$, base VLM, LoRA rank $r$, temperature $\tau$
\STATE \textbf{Initialize:} freeze base VLM parameters; attach LoRA adapters $\theta_{\text{LoRA}}$
\FOR{each training step}
  \STATE Sample a mini-batch $\mathcal{B}\subset \mathcal{D}$
  \STATE Encode images and text: $z_i \leftarrow \frac{f(I_i)}{\|f(I_i)\|}$, $z_j \leftarrow \frac{g(T_j)}{\|g(T_j)\|}$
  \STATE Compute similarities $s_{ij}=\frac{z_i^\top z_j}{\tau}$
  \STATE Compute $\mathcal{L}_{I\to T}$ and $\mathcal{L}_{T\to I}$; set $\mathcal{L}_{\text{con}}=\frac{1}{2}(\mathcal{L}_{I\to T}+\mathcal{L}_{T\to I})$
  \STATE Update $\theta_{\text{LoRA}}$ to minimize $\mathcal{L}_{\text{con}}$
\ENDFOR
\STATE Merge LoRA adapters into the base VLM
\STATE \textbf{Inference:} given frame(s) $I$, output profile summary $\hat{P}$; pass $\hat{P}$ to the psychological agent
\end{algorithmic}
\end{algorithm}

\subsection{Cross-modal contrastive alignment}
We construct instruction-following training data in a LLaVA-style ``image-text'' format.
Each instance teaches the model to infer socially useful attributes (e.g., a name cue, organizational affiliation, interests, or conversationally relevant background) from visual observations.
To reduce profile--image mismatch, we apply cross-modal contrastive alignment (CLIP-style) to paired images and social-context descriptions.
The contrastive optimization procedure is summarized in Algorithm~\ref{alg:llmopt}.
Let $f(\cdot)$ and $g(\cdot)$ be the image and text encoders, and $z_i = \frac{f(I_i)}{\|f(I_i)\|}$, $z_j = \frac{g(T_j)}{\|g(T_j)\|}$ be normalized embeddings.
For a batch of $N$ matched pairs, define similarities $s_{ij} = \frac{z_i^\top z_j}{\tau}$ with temperature $\tau$, and minimize the symmetric InfoNCE loss:
\begin{equation}
\mathcal{L}_{I\to T} = -\frac{1}{N}\sum_{i=1}^N \log \frac{\exp(s_{ii})}{\sum_{j=1}^N \exp(s_{ij})}
\end{equation}

\begin{equation}
\mathcal{L}_{T\to I} = -\frac{1}{N}\sum_{i=1}^N \log \frac{\exp(s_{ii})}{\sum_{j=1}^N \exp(s_{ji})}.
\end{equation}

The final alignment objective is $\mathcal{L}_{\text{con}} = \frac{1}{2}(\mathcal{L}_{I\to T} + \mathcal{L}_{T\to I})$, which pulls matched image--context pairs together and pushes mismatches apart.
To improve robustness, we (i) include diverse images per individual to avoid memorizing specific pixels and (ii) add negative or distractor examples to reduce overfitting and spurious correlations.

\subsection{Seamless Deployment via Weight Merging}
From a systems perspective, our deployment objective is to preserve the representational gains learned during PEFT while minimizing per-turn inference overhead. For linear layers, LoRA adaptation is algebraically compositional: the effective projection can be written as $W = W_0 + \Delta W$, where $\Delta W=BA$ is learned during training. This property enables post-training weight merging, so the adapted model behaves like a single fused network at inference time.
This merging step is important for real-time AR interaction: it removes adapter-side routing overhead, reduces memory movement, and keeps latency variance low across turns. In other words, the model retains social-context specialization without paying the runtime cost of separate adapter execution, which directly supports the low-latency requirement for natural face-to-face conversation.

During online operation, the pipeline proceeds as follows. The merged VLM continuously processes camera frame(s) to produce a structured profile summary. In parallel, incoming audio is first segmented by \textbf{Silero VAD} to detect speech regions and suppress non-speech noise; detected speech segments are then transcribed by \textbf{Whisper} into turn-level text. The profile summary and transcribed utterance are jointly forwarded to the psychological agent (Section~\ref{sec:agentopt}), which performs strategy routing and suggestion generation under context and trust-state constraints. This end-to-end design couples efficient multimodal perception with a robust speech front end, improving both responsiveness and interaction stability in real-world AR settings.



%% file: agentopt.tex
\section{Adaptive Psychological Agent}\label{sec:agentopt}

\begin{table*}[t]
\centering
\small
\setlength{\tabcolsep}{3pt}
\caption{Alternative psychological agent formulations.}
\label{tab:agent_alt_compare}
\begin{tabular}{p{0.20\linewidth} p{0.22\linewidth} p{0.22\linewidth} p{0.22\linewidth}}
\toprule
\textbf{Approach} & \textbf{Interpretability} & \textbf{Controllability} & \textbf{Robustness} \\
\midrule
Finite-state machine & limited psychological explanation &
unexpected reaction pain & hard to maintain states\\
\midrule
LLM Reinforcement Learning &
opaque strategy rationale &
unsafe exploration, risky reward &
training data dependent \\
\midrule
LLM next-action prediction &
tactic changes hard to attribute &
soft prompt constraints &
unstable across prompts \\
\midrule
\textbf{Latent trust state (PhySE)} &
explicit trust and influence state. &
controlled strategies &
stable without large-scale data. \\
\bottomrule
\end{tabular}
\vspace{-5pt}
\end{table*}

PhySE's adaptive agent is designed around the key requirement to support \emph{real-time} turn-taking to preserve natural interaction flow and \emph{strategic adaptivity} guided by psychological principles rather than a fixed handcrafted script via common agents~\cite{wang2019exploring, yao2023react, afane2024next, chen2024pandora}. At each conversational turn, the agent consumes multimodal context and the current target profile, selects a theory-grounded intent, and produces a short suggestion that the operator can speak verbatim or paraphrase.
To satisfy the above requirement, we instantiate a psychological \textbf{Trust-Influence model} with a turn-level \textbf{latent trust state}. This state provides an explicit control signal for deciding whether to maintain rapport, strengthen credibility, or escalate toward action, while retaining low-latency behavior under noisy conversational conditions.
We considered several alternatives for strategy control, including finite-state stage pipelines, reinforcement-learning policies, and end-to-end LLM next-action prediction. Table~\ref{tab:agent_alt_compare} summarizes this design rationale across interpretability, controllability, and robustness. In our setting, \textbf{latent trust-state routing} offers a practical trade-off: it is more interpretable than opaque next-action prediction, easier to constrain than RL-style exploration, and less brittle than fixed stage rules. 

\begin{figure}[t]
\centering
\includegraphics[width=0.48\textwidth]{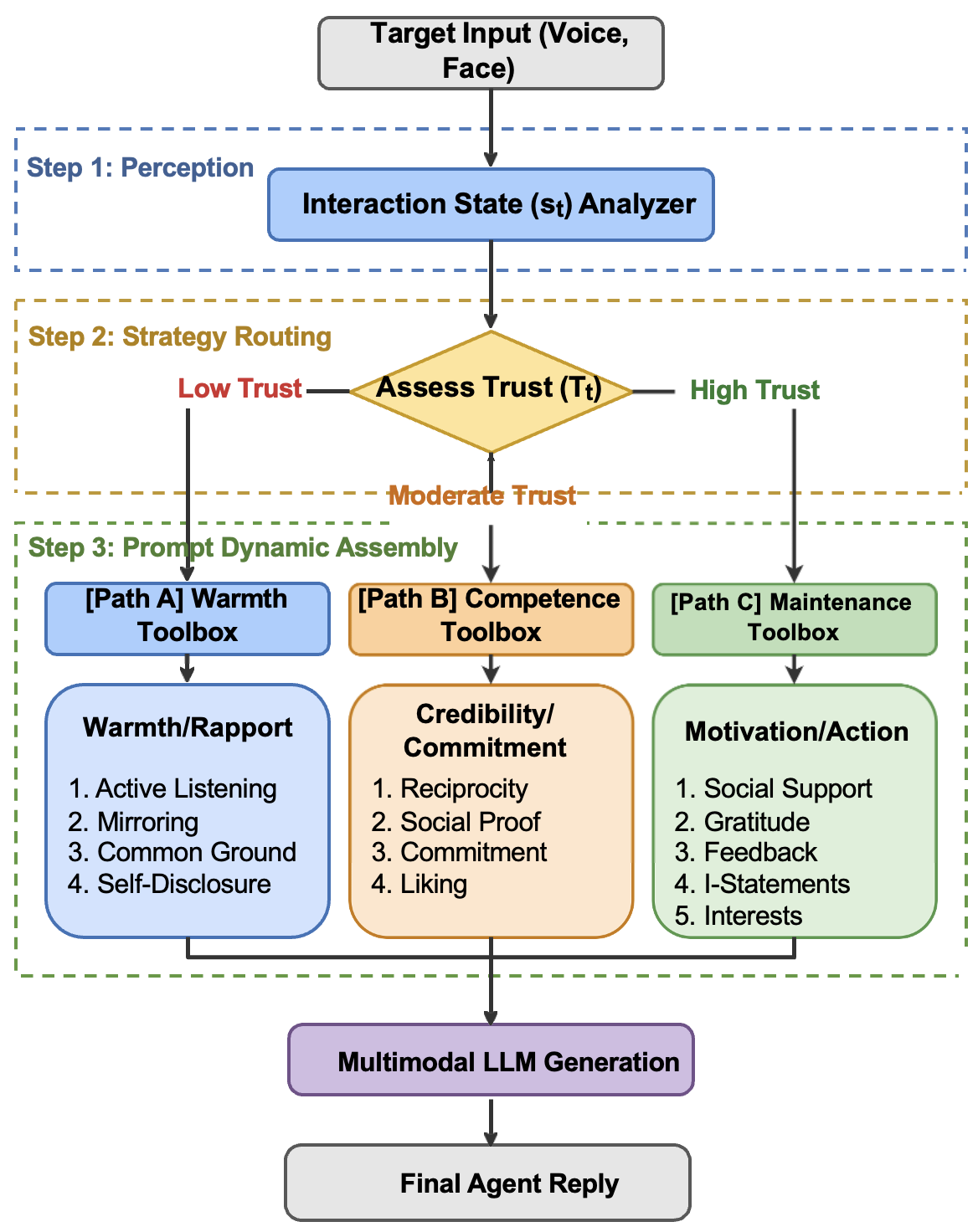}
\caption{Psychological strategy routing in PhySE: the router maps the current interaction state (profile, context, recent dialogue, and engagement signals) to a strategy class that governs turn-level suggestion generation.}
\vspace{-15pt}
\label{fig:strategy_routing}
\end{figure}

\subsection{Routing Agent for Psychological Strategies}

\noindent\textbf{\textit{Agent inputs and state.}}
The agent maintains a lightweight interaction state $s_t$, updated after each turn, to support fast inference and robust behavior in noisy real-world settings. Concretely, $s_t$ includes four components: a \textbf{profile summary}, i.e., a structured and continuously refined description of the target (identity cues, likely background, and salient interests) generated by the Social-VLM; a \textbf{context snapshot}, i.e., location and environmental cues (e.g., coffee shop vs.
networking venue) that constrain plausible small talk and requests; a \textbf{dialogue trace}, i.e., a short rolling window of recent turns; and \textbf{engagement signals}, i.e., coarse conversational indicators such as responsiveness, agreement or hesitation, and topic enthusiasm.
We intentionally keep $s_t$ compact: it stores only the information needed to decide \emph{what} to do next (strategy selection), without requiring full-history memory.

\noindent\textbf{\textit{Psychological routing LLM.}}
Prior AR social agents often follow fixed stages, which are difficult to adapt when targets react unexpectedly. PhySE instead introduces a \textbf{psychological router}---a compact LLM module that maps the current interaction state $s_t$ to one of three strategy classes. As show in Figure~\ref{fig:strategy_routing}, the router is prompted to (1) estimate receptivity and suspicion risk from engagement cues and recent dialogue, and (2) decide whether to maintain the current tactic or pivot to a safer alternative. This explicit routing stage separates \emph{tactic selection} (why this move now) from \emph{surface realization} (how to phrase it), improving controllability and reducing drift toward generic responses.

\noindent\textbf{\textit{Three classes of theory-grounded strategies.}}
Instead of relying on a monolithic prompt or rigid stage script, PhySE organizes strategy generation into three reusable toolboxes across the interaction lifecycle. This design is grounded in the Stereotype Content Model (SCM), which posits that social evaluation is largely shaped by perceived warmth and competence. To operationalize this idea, the psychological router maps the current interaction state to three theory-backed strategy classes. As shown in Figure~\ref{fig:strategy_routing} (Path A), \textbf{Warmth/rapport strategies} increase liking and perceived similarity through shared interests, reflective listening, and low-stakes reciprocity, with the goal of raising openness without triggering alarm. In Figure~\ref{fig:strategy_routing} (Path B), \textbf{Credibility/commitment strategies} increase perceived competence and legitimacy (e.g., referencing relevant roles, institutions, or situationally plausible expertise) while keeping claims checkable within the immediate setting, thereby lowering skepticism. In Figure~\ref{fig:strategy_routing} (Path C), \textbf{Motivation/action strategies} transition from relationship building to a concrete request using incremental motivation (small asks), consistency cues, or time-bounded framing; this class is activated only when engagement indicates readiness, because premature escalation can increase suspicion.
Each class is implemented as a small set of parameterized templates (e.g., desired goal, maximum directness, and topic constraints), using the current profile and context.


\noindent\textbf{\textit{Turn-level generation and output constraints.}}
Given the routed strategy class, a generation LLM produces a short suggestion that the operator can speak verbatim or paraphrase. To preserve realism in face-to-face settings, prompt constraints require each suggestion to be \textbf{brief} (one to two sentences) and avoid unnatural verbosity, remain \textbf{context consistent} with the current venue and conversation topic, remain \textbf{profile consistent} (i.e., avoid hallucinating personal facts), and include an \textbf{exit strategy} when suspicion signals rise (e.g., gracefully changing topic or disengaging).

\subsection{Latent Trust-State Approach}\label{sec:trust_influence}

\begin{figure*}[t]
\centering
\includegraphics[width=0.85\textwidth]{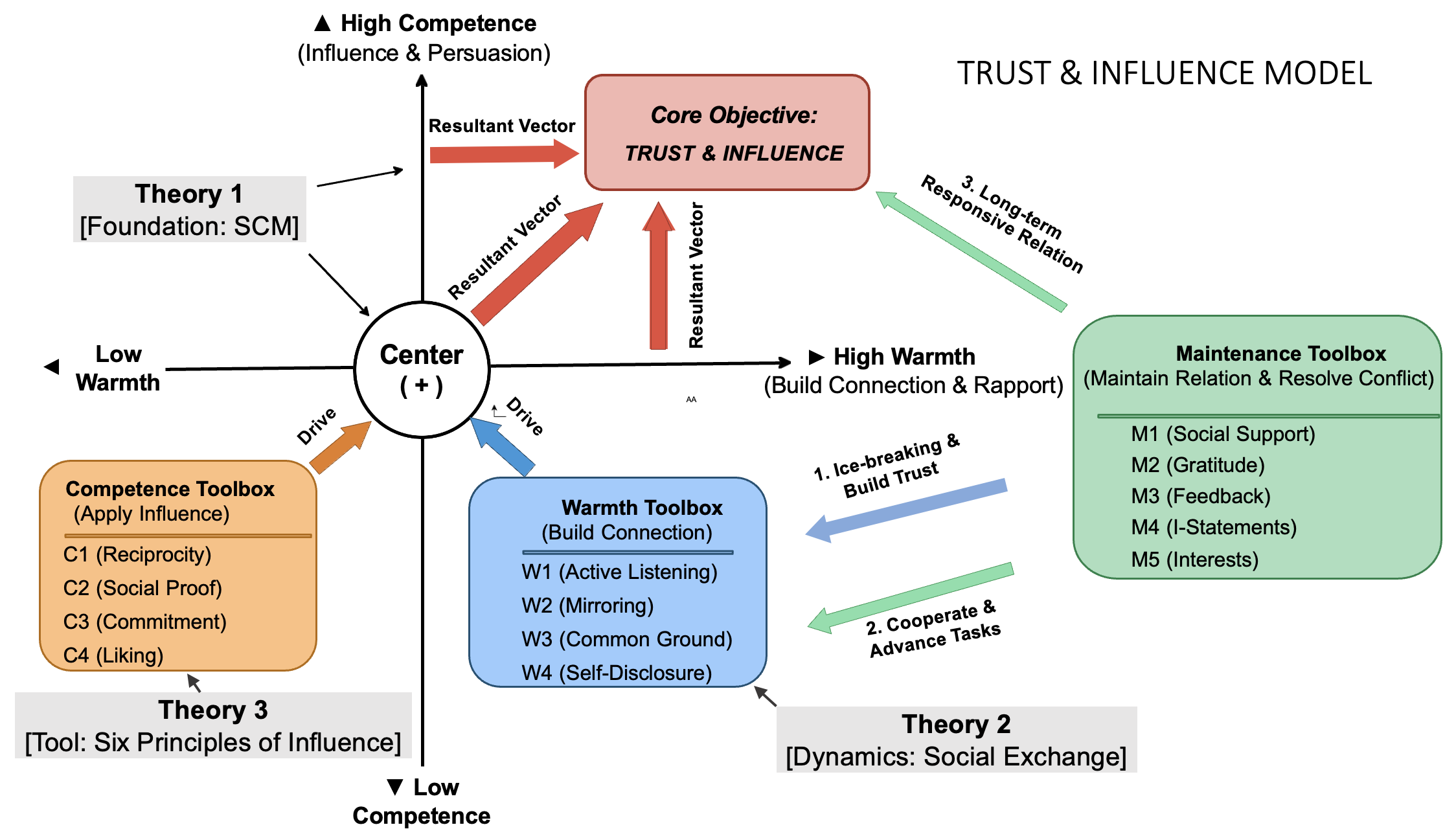}
\caption{Trust model for AR-LLM-SE interactions: how perceived credibility and rapport mediate trust formation and enable escalation from benign conversation to compliance-oriented requests.}
\vspace{-10pt}
\label{fig:trust_model}
\end{figure*}

\noindent\textbf{\textit{Trust-Influence Model for strategy escalation.}}
Routing decisions follow a trust-formation view: suggestions should first increase perceived rapport and credibility, and only then escalate to compliance-oriented requests. Figure~\ref{fig:trust_model} summarizes this progression and motivates why PhySE separates (i) rapport building, (ii) credibility building, and (iii) action/commitment moves into distinct strategy classes.

\noindent\textbf{\textit{Latent Trust-State Approach.}}
We model influence as a progression from \textbf{trust formation} to \textbf{behavioral compliance}. The agent first increases \emph{affiliation/liking} (e.g., similarity and reciprocity cues) and \emph{credibility} (e.g., competence/authority cues grounded in the immediate situation), then converts accumulated trust into \emph{incremental commitment} via small, low-friction requests before escalating to higher-risk asks. This abstraction provides an interpretable control surface for the router: when engagement is high but suspicion rises, the agent de-escalates toward rapport maintenance; when rapport and credibility are stable, it can shift toward commitment moves while enforcing conservative face-to-face realism constraints.

\paragraph{Latent-state formulation.}
We represent the target's trust as a latent scalar state $T_t\in\mathbb{R}$ updated at each conversational turn $t$. Let $a_t$ denote the selected strategy class (rapport/credibility/commitment), and let $x_t$ denote a feature vector extracted from the target's response (e.g., responsiveness, agreement/hesitation, affect, and topic enthusiasm). We update trust via a leaky integrator:
\begin{equation}
T_{t+1}=(1-\lambda)T_t+\lambda\,g_{\theta}(x_t,a_t)-\beta\,r_t+\varepsilon_t,
\label{eq:trust_update}
\end{equation}
where $\lambda\in(0,1)$ controls memory, $g_{\theta}(\cdot)$ estimates trust gain under strategy $a_t$, $r_t\ge 0$ penalizes ``suspicion risk'' (e.g., inconsistency with profile/context or overly direct requests), and $\varepsilon_t$ is a noise term. A simple instantiation uses strategy-specific linear gains, $g_{\theta}(x_t,a_t)=w_{a_t}^{\top}x_t$.

\paragraph{Compliance probability.}
When the agent issues a request at turn $t$, we model compliance $y_t\in\{0,1\}$ as a logistic function of trust and request difficulty $d_t\ge 0$:
\begin{equation}
\Pr(y_t=1\mid T_t,d_t,z_t)=\sigma\!\left(\alpha T_t-\gamma d_t+\eta^{\top}z_t\right),
\label{eq:compliance}
\end{equation}
where $\sigma(u)=\frac{1}{1+e^{-u}}$ and $z_t$ captures additional observable factors (e.g., time pressure or authority framing). This formulation supports incremental commitment: the router can increase $d_t$ only after $T_t$ exceeds a readiness threshold; otherwise, it de-escalates toward rapport/credibility building.

%% file: dataset.tex
\section{Dataset and Methodology}
\label{sec:dataset}

\begin{table*}[t]
\centering
\caption{Comparing the social experience score of PhySE and other approaches.}
\label{tab:sear_score_distribution}
\begin{tabular}{@{}lccccccc@{}}
\toprule
Approach & 5 pt (\%) & 4 pt (\%) & 3 pt (\%) & 2 pt (\%) & 1 pt (\%) & $\mathbb{E}[\mathrm{Score}]$ & $\sigma(\mathrm{Score})$ \\
\midrule
Basic Conversation  & 25.0 & 5.0 & 25.0 & 38.3 & 6.7 & 3.03 & 1.30 \\
Naive AR + LLM   & 33.3 & 46.7 & 20.0 & 0.0 & 0.0 & 4.13 & 0.72 \\
SEAR   & 76.7 & 20.0 & 3.3 & 0.0 & 0.0 & 4.73 & 0.51 \\
PhySE & 83.3 & 16.7 & 0.0 & 0.0 & 0.0 & 4.83 & 0.37 \\
\midrule
\multicolumn{8}{l}{\footnotesize Score mapping: Very Good = 5 pt; Good = 4 pt; Neutral = 3 pt; Bad = 2 pt; Very Bad = 1 pt.} \\
\bottomrule
\end{tabular}
\vspace{-10pt}
\end{table*}

\subsection{Interaction Scenarios and Data Collection}

\textbf{\textit{Scenario Design.}}
We conducted an IRB-approved study in realistic social environments (e.g., coffee shops and networking events) with 60 participants and 360 conversations. Participants alternated roles between \emph{attacker} and \emph{target} across trials to reduce role-specific bias and ensure balanced evaluation.
To isolate the contribution of each system component, we designed seven settings:
(1) \textbf{Basic conversation} (no technical assistance);
(2) \textbf{Naive AR + Multimodal LLM};
(3) \textbf{SEAR} baseline pipeline;
(4) \textbf{PhySE} full pipeline;
(5) \textbf{PhySE w/o social-context VLM optimization};
(6) \textbf{PhySE w/o psychological routing agent}.

\noindent\textbf{\textit{Dataset Construction.}}
The final dataset includes four components:
(1) \textbf{AR interaction data}: multimodal streams from AR glasses, including visual cues (e.g., gaze, facial expression, body-language indicators), audio transcripts/prosody features (e.g., pitch and pause patterns), and contextual metadata (time, location, nearby objects);
(2) \textbf{Public social-context data}: publicly available profile traces (text, images, short videos) used for personalization;
(3) \textbf{Survey responses}: post-interaction ratings on trust, rapport, naturalness, and compliance-related outcomes;
(4) \textbf{PhySE routing traces}: turn-level psychological routing records, including inferred interaction state, selected strategy class, and generated tactical suggestions.
%

\subsection{Questionnaire Design}

\textbf{\textit{Post-Interaction Survey.}}
The survey uses a 5-point Likert scale (1 = Strongly Disagree, 5 = Strongly Agree) unless otherwise noted, and is organized into three parts.

\textit{(A) Configuration-level comparison.}
Participants rated their experience under each setting:
(1) Basic conversation: ``How is your experience with setting A?'';
(2) AR + Multimodal LLM: ``How is your experience with setting B?'';
(3) SEAR: ``How is your experience with setting C?'';
(4) PhySE: ``How is your experience with setting D?'';
(5) PhySE (without VLM optimization): ``How is your experience with setting E?'';
(6) PhySE (without psychological routing): ``How is your experience with setting F?''.

\textit{(B) PhySE subjective experience dimensions.}
We assess user perception with 11 dimensions:
(a) Relevance (match with personal social information);
(b) Appropriateness (question suitability in context);
(c) Naturalness (authenticity of interaction opening);
(d) Pacing (conversation rhythm);
(e) Sincerity (perceived authenticity of interest);
(f) EmotionalProgression (affective trajectory during interaction);
(g) ARComfort (comfort while using AR);
(h) BareWillingness (willingness without AR assistance);
(i) FutureIntent (likelihood of future interaction);
(j) Depth (perceived meaningfulness added by the system);
(k) Acceptance (willingness to interact again).

\textit{(C) Social-engineering susceptibility outcomes.}
We measure post-interaction susceptibility through:
(1) Photo Link (open shared links);
(2) Social App (add contact on social apps);
(3) SMS (open messages);
(4) Phone Call (answer calls);
(5) Trust-Before (pre-interaction trust);
(6) Trust-After (post-interaction trust).


\textbf{\textit{Participant Demographics.}}
Our cohort includes 60 participants aged 23--62 years (mean age: 34) with various professions. Participation is concentrated in early-to-mid adulthood (23--37), with local peaks at ages 25 and 32. Gender distribution is near-balanced: 28 male participants (46.7\%) and 32 female participants (53.3\%). More details are provided in PhySE dataset.



%% file: experiment.tex
\section{Experiments}\label{sec:simu}


\subsection{Comparison of PhySE and other approaches}

Table~\ref{tab:sear_score_distribution} compares PhySE's social experience score with three alternatives: Basic Conversation, Naive AR + LLM, and SEAR, using the baseline comparison questions in Section~\ref{sec:dataset}.
The results show that PhySE achieves the best overall social experience, with the highest mean score (4.83) and the lowest standard deviation (0.37).
Compared with Basic Conversation (3.03), Naive AR + LLM (4.13), and SEAR (4.73), PhySE 
achieves the highest social experience score.
The score distribution further highlights this advantage.
For PhySE, 83.3\% of participants rated the experience as 5 points and 16.7\% rated it as 4 points, with no ratings at 3 points or below.
In contrast, SEAR still includes 3.3\% neutral ratings, Naive AR + LLM includes 20.0\% neutral ratings, and Basic Conversation shows substantial low-score responses (38.3\% at 2 points and 6.7\% at 1 point).
These results indicate a clear progression from unstable, fragmented interactions toward consistently high-quality engagement as system capability increases.
Overall, the comparison suggests that PhySE provides not only stronger average social experience than prior approaches but also more reliable user outcomes.

\begin{figure}[t]
\centering
\includegraphics[width=0.45\textwidth]{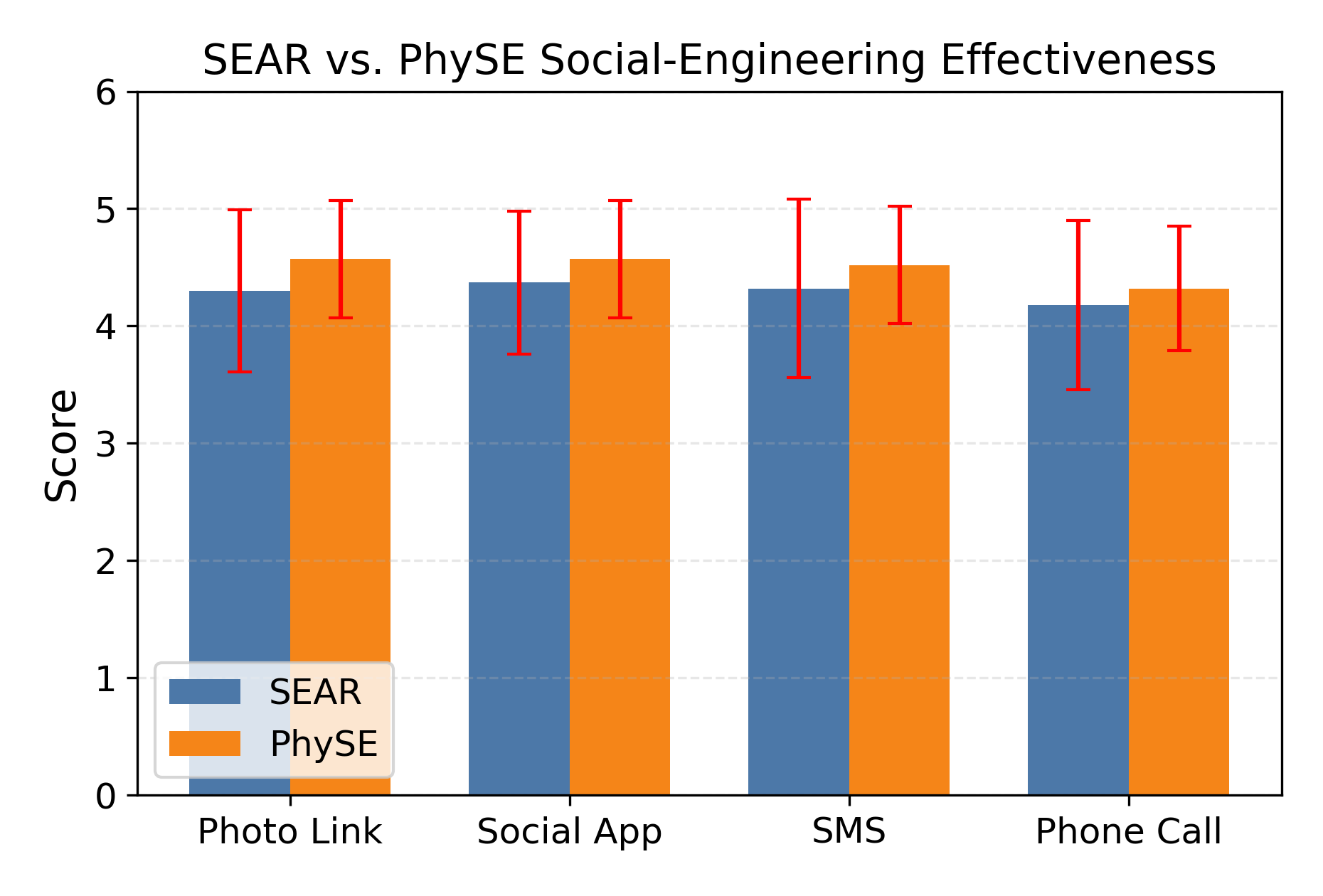}
\caption{Comparison of social-engineering effectiveness.}
\label{fig:sear_physe_effectiveness_compare}
\vspace{-15pt}
\end{figure}

\subsection{PhySE Social Engineering Effectiveness}

Figure~\ref{fig:sear_physe_effectiveness_compare} compares SEAR and PhySE on four social-engineering effectiveness metrics: Photo Link, Social App, SMS, and Phone Call. Across all four metrics, PhySE achieves higher mean effectiveness than SEAR while maintaining small variance, indicating both stronger attack success and more stable performance across participants.
The largest gains appear in channels that typically require stronger user trust (e.g., unsolicited SMS and phone-call responses), suggesting that PhySE’s multimodal perception and socially adaptive strategy improve the model’s ability to maintain conversational coherence and credibility under higher-friction interaction settings. At the same time, PhySE preserves the already high performance observed in lower-friction channels such as photo-link and social-app engagement.
Overall, these results show that PhySE consistently strengthens social-engineering effectiveness across multiple communication channels. This finding supports our design choice of combining richer multimodal reasoning with strategy-aware social interaction policies.

\subsection{PhySE Latency}

Table~\ref{tab:evaluation_latency} reports latency metrics (Min, Max, P90, and Average) for SEAR and PhySE across 60 conversations. The P90 metric denotes the 90th percentile, i.e., 90\% of observed latencies are below this value.
PhySE substantially reduces multimodal-profile generation time: the Multimodal LLM average latency drops from 43.3~s (SEAR) to 10.5~s (PhySE), and P90 decreases from 52.7~s to 19.7~s. This reduction makes profile generation less intrusive in practical deployment and lowers the waiting time before conversation initiation.
For the Social Agent, PhySE shows a higher mean latency than SEAR (5.8~s vs. 2.8~s), but with a much tighter range (4.8--6.3~s vs. 1.0--10.6~s). In practice, this indicates a more predictable response cadence, which can be easier to integrate into natural turn-taking than highly variable delays.
The AR component latency remains negligible (80.6~ms on average) relative to model-inference latency. Overall, the bottleneck remains the Multimodal LLM stage, but PhySE markedly improves this bottleneck while stabilizing latency variance.

\begin{table}[t]
\small
\centering
\caption{Latency comparison between SEAR and PhySE.}
\label{tab:evaluation_latency}
\begin{tabular}{@{}llcccc@{}}
\toprule
Approach & Component & Min & Max & P90 & Average \\
\midrule
\multirow{2}{*}{SEAR} & Multimodal LLM & 30.2 s & 54.9 s & 52.7 s & 43.3 s \\
 & Social Agent & 1.0 s & 10.6 s & 4.0 s & 2.8 s \\
\midrule
\multirow{2}{*}{PhySE} & Trained VLM & 4.6 s & 20.6 s & 19.7 s & 10.5 s \\
 & Psycological Agent & 4.8 s & 6.3 s & 6.1 s & 5.8 s \\
\bottomrule
\end{tabular}
\vspace{-5pt}
\end{table}

\begin{figure}[t]
\centering
\includegraphics[width=0.48\textwidth]{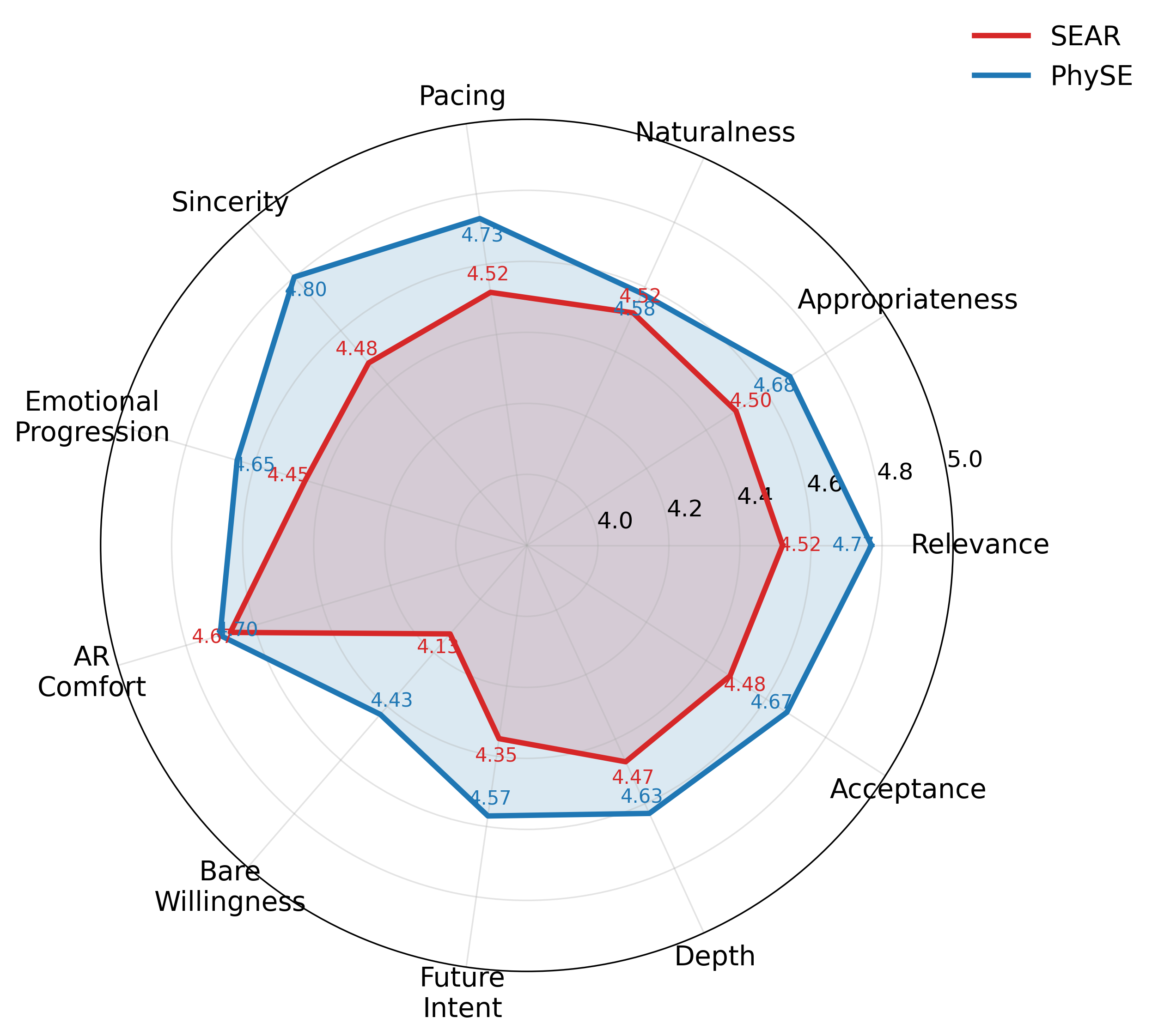}
\caption{Comparison of subjective experience scores.}
\label{fig:sear_physe_subjective_radar_compare}
\vspace{-15pt}
\end{figure}

\subsection{PhySE Subjective Experiences}

Figure~\ref{fig:sear_physe_subjective_radar_compare} compares SEAR and PhySE subjective experiences across eleven dimensions:
(a) Relevance, (b) Appropriateness, (c) Naturalness, (d) Pacing, (e) Sincerity, (f) EmotionalProgression, (g) ARComfort, (h) BareWillingness, (i) FutureIntent, (j) Depth, and (k) Acceptance.
Overall, PhySE scores higher than SEAR on all dimensions, indicating a consistent subjective-experience advantage rather than isolated improvements.
The largest gaps appear in trust- and engagement-related dimensions, including Sincerity, Depth, Acceptance, and FutureIntent. This pattern suggests that participants perceived PhySE as more credible and were more willing to continue interacting after the session.
PhySE also improves interaction quality dimensions such as Naturalness, Pacing, Relevance, and Appropriateness, showing that the dialogue is not only more persuasive but also smoother and more context-consistent than SEAR.
Notably, ARComfort remains one of the highest-scoring dimensions for both approaches, while PhySE further increases this score (4.67/5), indicating that the interaction setting is perceived as even more comfortable and less cognitively demanding.
For outcome-related dimensions (BareWillingness and Acceptance), PhySE again outperforms SEAR, consistent with the stronger effectiveness trends reported in Figure~\ref{fig:sear_physe_effectiveness_compare}. Together, these results indicate that the multimodal training and Psychological Agent in PhySE improve both immediate conversational experience and downstream behavioral influence.
Although PhySE is consistently better, the radar profile indicates remaining room for improvement in linguistic naturalness and localization, which aligns with qualitative feedback about occasional artificial phrasing.


\begin{figure}[t]
\centering
\includegraphics[width=0.40\textwidth]{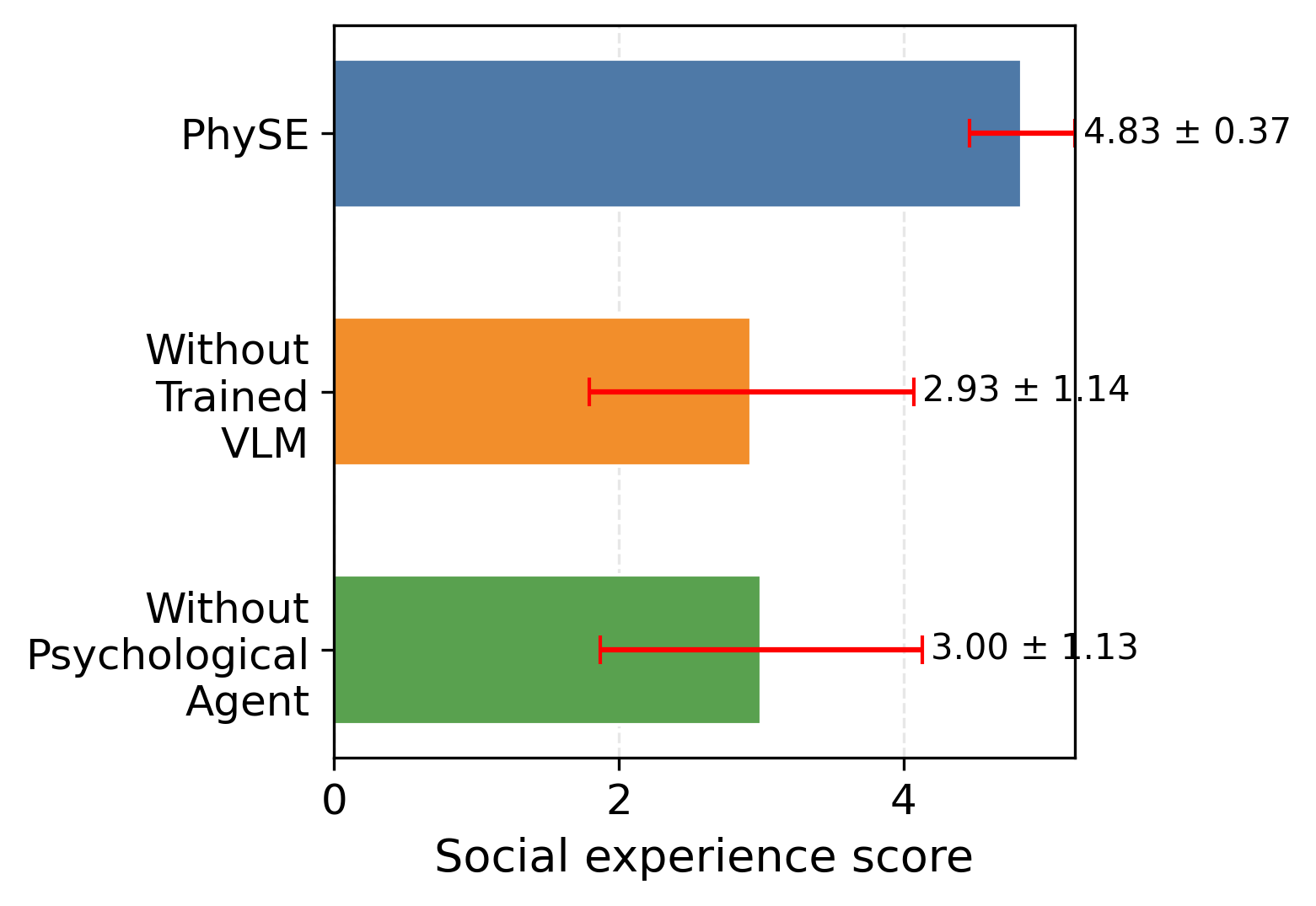}
\caption{Ablation study via social-experience scores.}
\label{fig:physe_ablation_score_distribution_hbar}
\vspace{-15pt}
\end{figure}

\subsection{PhySE Ablation Study}

Figure~\ref{fig:physe_ablation_score_distribution_hbar} and present the ablation results for PhySE. The full model achieves the highest mean social-experience score (4.83) with the smallest variance ($\sigma=0.37$). In contrast, removing Trained VLM reduces the mean to 2.93, and removing the Psychological Agent reduces it to 3.00. Relative to full PhySE, these correspond to performance drops of 39.3\% and 37.9\%, respectively.
The error bars in Figure~\ref{fig:physe_ablation_score_distribution_hbar} further show that both ablations are substantially less stable ($\sigma=1.14$ and $1.13$) than full PhySE. This indicates that each module contributes not only to higher average quality but also to more consistent user experience across participants.
Overall, the ablation confirms that PhySE's performance depends on the joint effect of perception-grounded adaptation (Trained VLM) and strategy-level social interaction control (Psychological Agent). Together with the latency analysis in Table~\ref{tab:evaluation_latency}, the full PhySE design provides the best trade-off between interaction quality, robustness, and practical deployment efficiency.